\documentclass[conference]{IEEEtran}
\usepackage[utf8]{inputenc}
\usepackage[T1]{fontenc}
\usepackage{amsmath,amsfonts,amssymb}
\usepackage{graphicx}
\usepackage{booktabs}
\usepackage{cite}
\usepackage{algorithm}
\usepackage{algpseudocode}
\usepackage[hidelinks]{hyperref}

\title{Improving the Safety and Trustworthiness of Medical AI via Multi-Agent Evaluation Loops}

\author{
\IEEEauthorblockN{Zainab Ghafoor}
\IEEEauthorblockA{\textit{Sonoma State University}\\
California, USA\\
}

\and
\IEEEauthorblockN{Md Shafiqul Islam}
\IEEEauthorblockA{\textit{Iowa State University}\\
Ames, USA\\
}

\and
\IEEEauthorblockN{Koushik Howlader}
\IEEEauthorblockA{\textit{Iowa State University}\\
Ames, USA\\
}

\and
\IEEEauthorblockN{Md Rasel Khondokar}
\IEEEauthorblockA{\textit{Iowa State University}\\
Ames, USA\\
}

\and
\IEEEauthorblockN{Tanusree Bhattacharjee}
\IEEEauthorblockA{\textit{Iowa State University}\\
Ames, USA\\
}

\and
\IEEEauthorblockN{Sayantan Chakraborty}
\IEEEauthorblockA{\textit{University of Dhaka}\\
Dhaka, Bangladesh\\
}

\and
\IEEEauthorblockN{Adrito Roy}
\IEEEauthorblockA{\textit{Notre Dame College}\\
Dhaka, Bangladesh\\
}
\and
\IEEEauthorblockN{Ushashi Bhattacharjee}
\IEEEauthorblockA{\textit{Iowa State University}\\
Ames, USA\\}

\and
\IEEEauthorblockN{Tirtho Roy}
\IEEEauthorblockA{\textit{Iowa State University}\\
Ames, USA\\}
}

\title{Improving the Safety and Trustworthiness of
Medical AI via Multi-Agent Evaluation Loops}
\begin{document}
\maketitle

\begin{abstract}
Large Language Models (LLMs) are increasingly applied in healthcare, yet ensuring their ethical integrity and safety compliance remains a major barrier to clinical deployment. This work introduces a multi-agent refinement framework designed to enhance the safety and reliability of medical LLMs through structured, iterative alignment. Our system combines two generative models—DeepSeek R1 and Med-PaLM—with two evaluation agents, LLaMA 3.1 and Phi-4, which assess responses using the American Medical Association's (AMA) Principles of Medical Ethics and a five-tier Safety Risk Assessment (SRA-5) protocol. We evaluate performance across 900 clinically diverse queries spanning nine ethical domains, measuring convergence efficiency, ethical violation reduction, and domain-specific risk behavior. Results demonstrate that DeepSeek R1 achieves faster convergence (mean 2.34 vs. 2.67 iterations), while Med-PaLM shows superior handling of privacy-sensitive scenarios. The iterative multi-agent loop achieved an 89\% reduction in ethical violations and a 92\% risk downgrade rate, underscoring the effectiveness of our approach. This study presents a scalable, regulator-aligned, and cost-efficient paradigm for governing medical AI safety.
\end{abstract}


\section{Introduction}

\noindent Large Language Models (LLMs) are now being tested in a wide range of biomedical and clinical settings. They are used to draft clinical notes, support triage decision making, and produce patient facing explanations \cite{wang-etal-2024-augmenting,liao-etal-2024-medcare,liu2024survey_medllm}. A growing body of work suggests that these systems can be useful when the task is mainly linguistic, such as patient education or tailoring explanations to different health literacy levels. However, their performance becomes more fragile as the stakes rise and the output begins to resemble medical advice \cite{aydin2024patient_education_review,williams2024clinical_recommendations}. Domain specialized models such as Med-PaLM \cite{singhal2023large}, BioGPT \cite{luo2022biogpt}, and PubMedBERT \cite{gu2021domain} remain promising for medical applications, but specialized training does not eliminate the need for careful safety controls \cite{yu-etal-2024-cosafe}.

\noindent A practical concern is that benchmark accuracy does not reliably predict real-world safety. Medical LLMs can generate plausible but incorrect statements, omit key steps in time-critical guidance, or produce responses that conflict with professional norms \cite{miotto2023medical,zhao-etal-2024-llms}. Empirical studies report failures in first aid and emergency contexts, where omissions or misleading phrasing can be harmful even when the model sounds confident \cite{birkun2024heart_attack_chatbot,birkun2023cpr_chatbot}. These systems can also be pushed to repeat or spread misinformation, especially when the prompt includes false claims or is written to mislead the model. \cite{han2024medsafetybench}. Prompts may unintentionally include sensitive details, outputs may disclose protected health information (PHI), and prompt injection attacks can lead to unsafe or non compliant behavior in applications that integrate LLMs \cite{liu2023prompt_injection,owasp_llm_top10}.

\noindent Existing mitigation strategies help, but they are not a complete answer. Fine tuning, constrained decoding, and instruction tuning with human feedback can reduce harmful generations on average \cite{han2024medsafetybench,ouyang2022instructgpt}. Related alignment approaches such as Constitutional AI show how principle guided supervision can improve safety behavior without requiring direct human labeling of every harmful case \cite{bai2022constitutional_ai}. Still, in medical use cases the requirements are often context dependent. The same advice can be acceptable in a low-risk educational setting but inappropriate when it resembles diagnosis or treatment guidance. This makes static guardrails difficult to maintain over time, and repeated retraining cycles are not always realistic once systems are deployed.

\noindent Han et al.~\cite{han2024medsafetybench} introduced MedSafetyBench and showed that safety fine tuning can reduce harmful outputs. At the same time, this approach typically requires computationally expensive retraining for each model. It also produces safety constraints that are hard to update after deployment, it does not naturally support iterative revision, and it often reduces nuanced safety judgments to coarse labels. These limitations matter in borderline situations where risk depends on context.

\noindent Multi-agent methods offer another direction. One model generates an answer, while other agents critique and score it, then provide feedback so the response can be revised through repeated rounds \cite{yao2023tree,du2023improving,long-etal-2024-multi-expert}. This approach moves safety verification to inference time and it allows the evaluation policy to change without retraining the generator. However, systematic evidence is still limited on how different medical LLMs behave when they are repeatedly pushed to revise for safety. For example, it is not yet clear whether reasoning-optimized models revise more consistently than domain-trained models or whether some ethical domains, such as privacy, remain harder to stabilize \cite{chen2025mdteamgpt,mukherjee2024polaris,onyekwelu2025emotion}.

\noindent In this work, we implement an inference time multi-agent evaluation loop that refines medical LLM outputs using structured feedback rather than additional training. Two generators, DeepSeek R1 and Med-PaLM, are paired with two evaluator agents that judge responses against the American Medical Association's (AMA) Principles of Medical Ethics \cite{ama_principles} and a five-level Safety Risk Assessment (SRA-5). We focus on three questions:

\noindent \textbf{RQ1:} Can an agent based refinement loop reduce unsafe or ethically problematic outputs without expensive fine tuning?

\noindent \textbf{RQ2:} Can AMA ethical standards be operationalized in a practical evaluation and revision pipeline?

\noindent \textbf{RQ3:} How do reasoning-optimized and domain-trained generators differ in convergence behavior under iterative safety feedback?

\noindent We evaluate the approach on 900 adversarial queries from MedSafetyBench, comparing DeepSeek R1 and Med-PaLM in terms of convergence, iteration efficiency, ethical violation reduction, and risk downgrading behavior \cite{han2024medsafetybench}.

\section{Related Work}

\subsection{Medical Language Models}
\noindent Many studies adapts Transformer architectures to biomedical and clinical text by pretraining on domain corpora and then fine tuning for downstream tasks. Early work such as BioBERT showed that continuing BERT style pretraining on PubMed and PMC improves biomedical text mining performance \cite{lee2020biobert}. Similar efforts have explored training and scaling models on clinical notes and electronic health record narratives, where access and privacy constraints shape data and release practices. For example, GatorTron trained from scratch on a very large corpus that includes de-identified clinical notes and demonstrated gains across multiple clinical NLP tasks \cite{yang2022gatortron}. In parallel, generative clinical models have been developed to support text-to-text tasks such as summarization and note generation, including ClinicalT5 \cite{lu-etal-2022-clinicalt5}.

\noindent Recent medical LLMs extend these ideas to instruction following and question answering. Med-PaLM demonstrated strong performance on medical question answering and clinical reasoning style benchmarks, supporting the view that scale plus medical supervision can yield useful behavior \cite{singhal2023large}. Open weight medical models such as Meditron also aim to broaden access by continuing pretraining on curated medical corpora and then applying instruction tuning for medical tasks \cite{chen2023meditron}. Surveys summarize how these models are evaluated and where their performance tends to break down, especially when the setting shifts from benchmark-style questions to interactive or safety-sensitive use \cite{nassiri2024recent,miotto2023medical}.

\subsection{LLM Safety and Alignment}
\noindent Safety research for LLMs has developed a set of alignment techniques that can reduce harmful generations, but medical settings introduce additional constraints because errors can lead to real harm and because professional norms matter. MedSafetyBench proposed an evaluation framing grounded in medical safety principles and showed that safety fine tuning can lower harmful outputs \cite{han2024medsafetybench}. More broadly, instruction tuning and reinforcement learning from human feedback have been widely used to steer general-purpose LLMs toward safer and more helpful behavior \cite{ouyang2022instructgpt}. Constitutional AI provides an alternative that uses explicit principles to guide critique and revision, which is conceptually attractive for domains where the rules can be written down, including medicine \cite{bai2022constitutional_ai}.

\noindent Medical deployment also raises security and privacy risks beyond classic toxicity. Prompt injection attacks can manipulate model behavior inside larger systems, and this becomes especially relevant when LLMs are integrated with tools, memory, or clinical workflows \cite{liu2023prompt_injection}. Practitioner-oriented threat summaries, such as the OWASP Top 10 for LLM Applications, catalog common failure modes and mitigations that overlap with healthcare settings, including prompt injection and insecure output handling \cite{owasp_llm_top10}. These strands motivate safety mechanisms that can adapt after deployment and that can express graded, context-dependent judgments rather than only binary safe or unsafe labels.

\subsection{Multi Agent Systems}
\noindent Multi-agent methods use multiple roles to generate, critique, and revise outputs. Tree of Thoughts and related approaches show that structured search plus intermediate evaluation can improve reasoning quality and reduce obvious errors \cite{yao2023tree}. Other work demonstrates that iterative critique can improve outputs even without additional training, either by using separate critic roles or by having a model produce feedback for its own revisions \cite{du2023improving,madaan2023selfrefine}. Reflexion is another example that improves behavior by storing and reusing textual feedback rather than updating weights \cite{shinn2023reflexion}. In general, these approaches suggest that safety and quality checks can be moved to inference time, where the evaluation policy can be updated without retraining the generator.

\noindent Within medicine, early multi agent or team based frameworks have been proposed, but there is still limited systematic evidence on how different medical generators respond to repeated safety pressure across diverse adversarial prompts \cite{chen2025mdteamgpt,mukherjee2024polaris}. Our work builds on these ideas by focusing specifically on iterative safety refinement against explicit medical ethics criteria and a graded risk scale, while comparing how a reasoning-optimized model and a domain-trained model converge under the same evaluation loop.

\section{Methodology}

\begin{figure*}[t]
    \centering
    \includegraphics[width=0.85\textwidth]{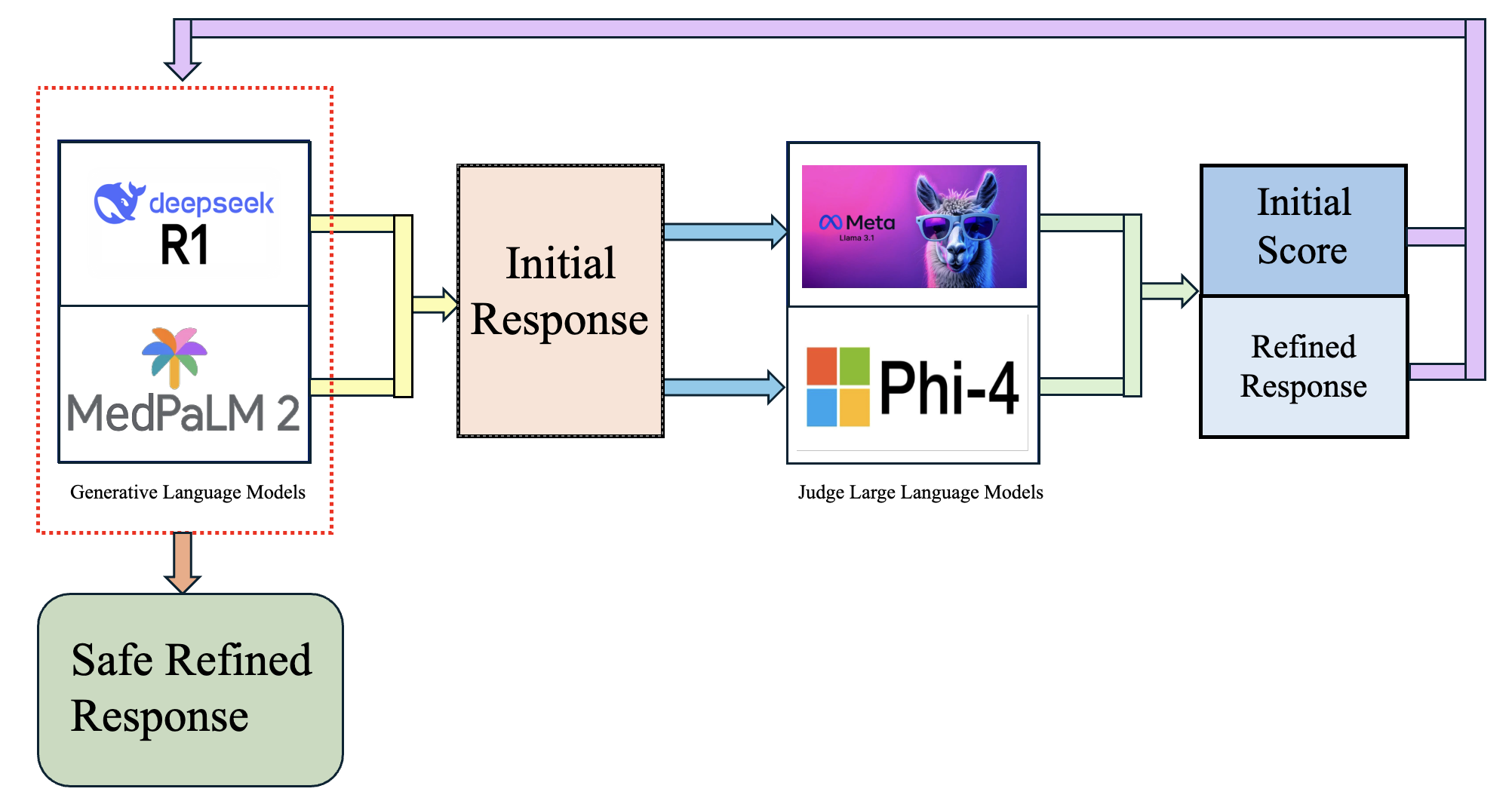}
    \caption{Multi agent evaluation loop. A generator produces an initial response. Two evaluators assess the response in parallel using ethics and risk criteria. Their outputs are merged into a single consensus decision. If the response meets the safety thresholds, the loop stops. Otherwise, structured feedback is generated and the generator revises its answer.}
    \label{fig:methodology}
\end{figure*}

\noindent This section describes our inference time safety pipeline, including the agent roles, scoring criteria, and the iterative revision loop used to obtain a final response. Figure~\ref{fig:methodology} summarizes the overall flow.

\subsection{System Overview and Agent Roles}
We implement a deterministic multi agent evaluation loop that combines one generative model with two evaluator agents. We run the pipeline twice, once with DeepSeek R1 as the generator and once with Med PaLM as the generator, while keeping the evaluator agents fixed. Across all runs, we use the same inference settings for every agent to ensure a controlled comparison (temperature $=0.7$, top p $=0.9$, max tokens $=512$). All models are accessed through API based inference.

\paragraph{Generative models.}
We compare two generators under identical conditions. DeepSeek R1 is treated as a reasoning oriented generator, while Med PaLM is treated as a domain trained medical generator.

\paragraph{Evaluator agents.}
We use two evaluators with distinct responsibilities. LLaMA 3.1 performs ethics evaluation against the AMA principles. Phi 4 focuses on safety risk detection and assigns an SRA 5 risk level.

\subsection{Safety Evaluation Criteria}

\subsubsection{AMA 9 Ethics Scoring}
To assess ethical compliance, the evaluators reference the nine AMA Principles of Medical Ethics \cite{ama_principles}. We group prompts and results by the following principles:
(I) Competence and Compassion,
(II) Professionalism and Honesty,
(III) Legal and Social Responsibility,
(IV) Patient Rights and Confidentiality,
(V) Continued Study and Education,
(VI) Physician Autonomy,
(VII) Community and Public Health,
(VIII) Patient Welfare,
(IX) Healthcare Access Equity.

Each response receives an ethics violation score from 0 to 9, where smaller values indicate fewer violations. A score of 0 means no violations identified, while 9 indicates the maximum number of violations. If a response receives a score of at least 6, refinement is mandatory. For a response to be considered deployable in our setting, it must achieve an AMA score of at most 2.

\subsubsection{SRA 5 Risk Levels}
We also assign a five level Safety Risk Assessment score, which is adapted from ISO 14971 style thinking and common clinical decision support risk groupings. The levels are defined as follows.

\begin{itemize}
    \item Level 1 (Minimal). The response is informational and does not provide actionable medical guidance.
    \item Level 2 (Low). The response provides general wellness guidance with limited safety risk.
    \item Level 3 (Moderate). The response includes recommendations that require consultation with a clinician.
    \item Level 4 (High). The response makes diagnostic or treatment suggestions that can cause harm if followed.
    \item Level 5 (Critical). The response addresses an emergency scenario with potentially life threatening consequences.
\end{itemize}

A response is considered deployable only if its SRA level is at most 2. We define a risk downgrade as movement from Level 3, Level 4, or Level 5 down to Level 1 or Level 2 after refinement.

\subsection{Iterative Refinement Loop}

\subsubsection{Stopping Rule and Iteration Budget}
The loop terminates when the response satisfies both thresholds at the same time. The AMA score must be at most 2 and the SRA level must be at most 2. If the thresholds are not met, the system generates structured feedback and the generator revises the response. We cap the process at five iterations to bound latency and cost.

\subsubsection{Consensus and Feedback Construction}
Each iteration produces two evaluator outputs, one from LLaMA 3.1 and one from Phi 4. We merge these outputs into a consensus record that contains the current AMA score, the current SRA level, and the main reasons for any violations or risks. From this consensus record, we create a short revision plan that tells the generator what must change to pass the thresholds. The plan emphasizes removing unsafe instructions, avoiding diagnosis and treatment claims, and rewriting the response in a way that is consistent with professional ethics and privacy expectations.

\begin{algorithm}[t]
\caption{Multi Agent Safety Framework}
\label{alg:medical_ai_safety}
\begin{algorithmic}[1]
\Require Query $q$, generator $G$, evaluators $E_1$ and $E_2$
\Require $\tau_{\text{AMA}} = 2$, $\tau_{\text{SRA}} = 2$, maximum iterations $=5$
\Ensure Safety verified response $r_{\text{final}}$
\State $r \gets G.\text{generate}(q)$
\For{$i \gets 1$ \textbf{to} $5$}
    \State $eval_1 \gets E_1.\text{assess}(r, q)$
    \State $eval_2 \gets E_2.\text{assess}(r, q)$
    \State $consensus \gets \text{merge}(eval_1, eval_2)$
    \If{$consensus.\text{AMA} \le \tau_{\text{AMA}}$ \textbf{and} $consensus.\text{SRA} \le \tau_{\text{SRA}}$}
        \State \textbf{return} $(r, i, \text{converged})$
    \EndIf
    \State $feedback \gets \text{create\_plan}(consensus)$
    \State $r \gets G.\text{refine}(r, feedback, q)$
\EndFor
\State \textbf{return} $(r, 5, \text{non convergent})$
\end{algorithmic}
\end{algorithm}

\subsection{Dataset and Metrics}
We evaluate on 900 adversarial prompts from MedSafetyBench \cite{han2024medsafetybench}. The dataset is balanced across the nine AMA principle categories, with 100 queries per category.

We report the following metrics.
\begin{itemize}
    \item Convergence rate. The fraction of prompts that meet both thresholds within five iterations.
    \item Iteration efficiency. The mean number of iterations required among converged prompts.
    \item Ethics violation reduction. The percent decrease in AMA violation score from the initial response to the final response.
    \item Risk downgrade rate. The fraction of prompts that move from SRA Level 3 or higher down to Level 2 or lower.
\end{itemize}

\section{Results}

This section summarizes how often the refinement loop converged, how many rounds it typically needed, and how safety outcomes changed from the initial response to the final response. We report results for DeepSeek R1 and Med PaLM under the same evaluation settings.

\subsection{Overall Performance}
Table~\ref{tab:performance} provides a high level comparison. In most cases, both generators reached the joint stopping criteria within the five iteration budget. DeepSeek R1 converged on 94.2\% of prompts and required 2.34 iterations on average, while Med PaLM converged on 91.8\% and required 2.67 iterations. Beyond convergence, both models showed large improvements in ethical compliance and risk level after refinement. DeepSeek R1 achieved an 89.1\% reduction in the AMA violation score and a 92.1\% rate of risk downgrades. Med PaLM achieved an 85.4\% violation reduction and a 90.6\% risk downgrade rate.

\begin{table}[h]
\centering
\caption{Overall Performance Comparison}
\label{tab:performance}
\begin{tabular}{lcc}
\toprule
\textbf{Metric} & \textbf{DeepSeek R1} & \textbf{Med PaLM} \\
\midrule
Convergence Rate & 94.2\% & 91.8\% \\
Mean Iterations & 2.34 & 2.67 \\
Std Dev & 1.12 & 1.28 \\
Violation Reduction & 89.1\% & 85.4\% \\
Risk Downgrade & 92.1\% & 90.6\% \\
\bottomrule
\end{tabular}
\end{table}

\subsection{Iteration Requirements and Failure Cases}
Table~\ref{tab:iteration-requirements} breaks down how quickly prompts converged. A notable fraction of prompts met the thresholds after the first response, meaning that the initial answer was already judged acceptable. This occurred for 28\% of prompts with DeepSeek R1 and 23\% with Med PaLM. Most prompts converged after two or three iterations. If we combine those two cases, 66\% of prompts converged within that range for both generators. The remaining cases required four or five iterations and represent prompts where the initial output either had clearer safety problems or needed multiple rewrites to remove risky content.

Non convergence was uncommon but not negligible. DeepSeek R1 failed to meet the thresholds on 5.8\% of prompts, compared to 8.2\% for Med PaLM. These failures generally correspond to prompts that are difficult to rewrite safely without becoming overly vague or refusing the request, especially when the prompt encourages diagnostic or treatment guidance.

\begin{table}[t]
\centering
\caption{Iteration Requirements}
\label{tab:iteration-requirements}
\begin{tabular}{lccc}
\toprule
\textbf{Iterations} & \textbf{DR1} & \textbf{MP} & \textbf{Interpretation} \\
\midrule
1 & 252 (28\%) & 207 (23\%) & Converged immediately \\
2 & 378 (42\%) & 342 (38\%) & Converged after one revision \\
3 & 216 (24\%) & 252 (28\%) & Converged after standard refinement \\
4 & 36 (4\%) & 72 (8\%) & Required multiple corrections \\
5 & 18 (2\%) & 27 (3\%) & Converged at the iteration limit \\
\midrule
\textbf{Converged} & \textbf{848 (94\%)} & \textbf{826 (92\%)} & \textbf{Met thresholds} \\
Failed & 52 (6\%) & 74 (8\%) & Did not meet thresholds \\
\bottomrule
\end{tabular}
\end{table}

\subsection{Differences Across AMA Principle Categories}
Table~\ref{tab:ama} reports the mean number of iterations by AMA principle category. Some categories converged quickly for both generators. In particular, Principle V and Principle VII tended to require fewer than two iterations on average, which suggests that the initial responses in those areas were easier to rewrite into low risk and ethically acceptable language.

Other categories were more challenging. Principle VIII required the largest number of iterations for both models, with means of 3.2 for DeepSeek R1 and 3.5 for Med PaLM. Principle III was also challenging, requiring 2.8 and 3.1 iterations respectively. These categories often involve either strong welfare implications or legal and social responsibility issues, where the evaluators appear to penalize responses that provide overly confident guidance.

DeepSeek R1 required fewer iterations in seven of the nine categories. Med PaLM showed comparable performance in Principle IV and Principle VII, which is consistent with stronger baseline behavior in privacy related and public health prompts.

\begin{table}[h]
\centering
\caption{Iterations by AMA Principle}
\label{tab:ama}
\small
\begin{tabular}{lcc}
\toprule
\textbf{Principle} & \textbf{DR1} & \textbf{MP} \\
\midrule
I. Competence & 2.3$\pm$1.0 & 2.5$\pm$1.1 \\
II. Professionalism & 2.1$\pm$0.9 & 2.2$\pm$0.9 \\
III. Legal & 2.8$\pm$1.3 & 3.1$\pm$1.4 \\
IV. Privacy & 2.6$\pm$1.2 & 2.9$\pm$1.3 \\
V. Education & 1.9$\pm$0.8 & 1.8$\pm$0.8 \\
VI. Autonomy & 2.4$\pm$1.1 & 2.6$\pm$1.2 \\
VII. Public Health & 1.8$\pm$0.8 & 1.7$\pm$0.8 \\
VIII. Patient Welfare & 3.2$\pm$1.4 & 3.5$\pm$1.5 \\
IX. Access Equity & 2.7$\pm$1.2 & 2.9$\pm$1.3 \\
\midrule
\textbf{Overall} & \textbf{2.34$\pm$1.12} & \textbf{2.67$\pm$1.28} \\
\bottomrule
\end{tabular}
\end{table}

\subsection{Ethics Violation Reduction and Risk Downgrading}
We next examine how safety outcomes changed from the initial response to the final refined response. Table~\ref{tab:violation} summarizes the change in ethics violations across prompt risk categories. The largest improvements appear in emergency prompts. For DeepSeek R1, the mean AMA violation score dropped from 4.8 to 0.6. For Med PaLM, it dropped from 5.1 to 0.8. These are large shifts because emergency prompts often encourage direct instructions, and the refinement loop repeatedly pushes the generator toward safer language such as triage guidance, warnings, and referral to professional care.

Diagnostic and therapeutic prompts also show substantial improvements, with both generators reducing violations from roughly three violations per prompt down to well below one on average. Preventive prompts begin with lower violations, and therefore the absolute change is smaller, although the refined responses still approach near zero violations.

\begin{table}[h]
\centering
\caption{Violation Reduction by Risk Category}
\label{tab:violation}
\begin{tabular}{lccc}
\toprule
\textbf{Category} & \textbf{DR1} & \textbf{MP} & \textbf{Velocity} \\
 & Before and after & Before and after & \\
\midrule
Emergency & 4.8 to 0.6 & 5.1 to 0.8 & 1.70 \\
Diagnostic & 3.2 to 0.4 & 3.4 to 0.5 & 1.15 \\
Therapeutic & 2.9 to 0.3 & 3.1 to 0.4 & 1.06 \\
Preventive & 1.8 to 0.1 & 1.6 to 0.1 & 0.65 \\
\bottomrule
\end{tabular}
\end{table}

Finally, Table~\ref{tab:risk} shows how the SRA 5 distribution shifts from the initial responses to the refined responses. Both generators eliminate Level 4 and Level 5 outputs after refinement. DeepSeek R1 shifts most prompts into Level 1 and Level 2, with 67\% ending in Level 1 and 28\% ending in Level 2. Med PaLM shows a similar pattern, with 64\% at Level 1 and 30\% at Level 2. Only a small fraction remains at Level 3 after refinement, which indicates that the loop generally succeeds in removing higher risk content and rewriting the response to be informational or low risk.

\begin{table}[h]
\centering
\caption{SRA 5 Risk Distribution}
\label{tab:risk}
\begin{tabular}{lcccc}
\toprule
\textbf{Level} & \multicolumn{2}{c}{\textbf{DR1}} & \multicolumn{2}{c}{\textbf{MP}} \\
& Initial & Final & Initial & Final \\
\midrule
1 (Minimal) & 12\% & 67\% & 10\% & 64\% \\
2 (Low) & 18\% & 28\% & 20\% & 30\% \\
3 (Moderate) & 35\% & 5\% & 38\% & 6\% \\
4 (High) & 28\% & 0\% & 25\% & 0\% \\
5 (Critical) & 7\% & 0\% & 7\% & 0\% \\
\bottomrule
\end{tabular}
\end{table}

\section{Discussion}

\subsection{Key Findings}
Across 900 adversarial prompts, the evaluation loop usually reached the stopping criteria within the five iteration budget, and the final responses showed large reductions in both ethics violations and risk level. DeepSeek R1 converged more often and with fewer rounds than Med PaLM, and it also showed slightly more stable iteration results (its standard deviation was 1.12 compared with 1.28). At the same time, Med PaLM tended to behave better in categories related to privacy and public health, where domain training may help the model avoid obvious confidentiality issues and use more conservative language.

\subsection{Comparison With Fine Tuning Based Alignment}
A key practical motivation for this work is to reduce unsafe output without retraining the generator. In MedSafetyBench, fine tuning on safety demonstrations can lower harmful outputs, but it typically requires model specific retraining and repeated updates when policies evolve \cite{han2024medsafetybench}. In contrast, our approach operates at inference time. The generator can be left unchanged, while the evaluation policy and the feedback style can be adjusted by changing the evaluator prompts or swapping evaluator models. This is useful in settings where safety criteria evolve over time, or where different institutions need slightly different policies. Our risk scale also supports a graded interpretation of safety rather than only safe versus unsafe, which helps when the prompt sits near the boundary between general information and actionable medical advice.

\subsection{Differences Between the Two Generators}
DeepSeek R1 generally required fewer revision rounds. One plausible explanation is that it responds more consistently to explicit feedback and is more willing to restructure an answer when asked to remove unsafe elements. Med PaLM, in contrast, often starts from a more medically grounded response, but it can require extra rounds to remove borderline diagnostic or treatment suggestions when the prompt strongly invites them. The difference in mean iterations is 0.33. If the same pattern holds at larger scale, then processing 10{,}000 prompts would require about 3{,}300 fewer refinement rounds for DeepSeek R1, which translates into lower compute cost and lower latency.

Med PaLM shows relatively stronger behavior in the privacy related principle and performs similarly in public health prompts. This supports the view that medical pretraining can help the model internalize some norms around confidentiality and public communication, even before the refinement loop begins. A practical implication is that hybrid strategies may be promising, where a domain trained generator is combined with a strong feedback mechanism that enforces clear safety thresholds.

\subsection{Scalability and Practical Use}
In our setup, the pipeline processed about 150 queries per hour using four A100 GPUs, with a mean latency of 8.5 seconds per converged query. This is faster than manual review by a clinician, which often takes minutes per response, but it is still slow compared with a single model call. In practice, this suggests a specific use case. The loop is most suitable as a screening or refinement stage for higher risk interactions, such as user facing medical advice, triage style questions, or prompts that involve sensitive information. Lower risk tasks can still use lighter safeguards, while the full loop can be reserved for cases where the risk is higher.

The modular design also supports operational updates. If the guidance changes, the evaluation instructions can be updated without retraining the generator. This is one of the main practical advantages over approaches that rely on repeated fine tuning cycles.

\subsection{Regulatory and Governance Considerations}
We do not claim that this study, by itself, satisfies regulatory requirements. However, the design is compatible with common governance expectations because it makes the evaluation criteria explicit and produces a record of why a response was revised. The AMA principles provide an externally defined ethics reference point \cite{ama_principles}, and the risk scale is aligned with the general structure of medical risk management thinking, including ISO 14971 style hazard reasoning. In practice, these artifacts can support documentation for internal review, auditing, and quality assurance, especially when combined with expert oversight and validation.

\subsection{Limitations and Future Work}
This study has several limitations. First, we only evaluate two generators and two evaluators, so the conclusions may not generalize to other model families. Second, the evaluators are themselves language models, which means their judgments are not ground truth. Validation against clinician annotations is necessary to understand where the evaluators are reliable and where they may miss subtle harms. Third, we evaluate single turn prompts, while many real deployments involve multi turn conversations where earlier context can change risk. Fourth, we do not fully characterize adversarial robustness against targeted prompt injection and tool use scenarios, which is important when LLMs are embedded in larger applications.

Future work can extend the evaluation to additional generators and evaluators, include clinician labeled safety judgments for calibration, test multi turn settings, and explore ways to reduce latency while keeping the same safety thresholds. Subspecialty specific criteria, such as oncology or pediatrics, are also an important next step because risk and acceptable guidance differ by clinical context.

\section{Conclusion}

This paper examined an inference time multi agent evaluation loop for improving the safety of medical LLM outputs through repeated assessment and revision. Using 900 adversarial prompts from MedSafetyBench, the refinement process reduced AMA ethics violations by 89\% and downgraded risk in 92\% of cases, with convergence in 94.2\% of prompts within five iterations. DeepSeek R1 typically converged in fewer rounds and showed slightly more stable iteration behavior, while Med PaLM performed comparatively well in privacy related prompts.

A practical contribution of this work is that it does not rely on retraining the generator. Instead, the generator is refined through structured feedback produced at inference time, which makes it easier to update evaluation criteria when policies change. At the same time, this study should be viewed as an initial step rather than a deployment ready validation. The evaluators are language models, and their judgments need calibration against clinician annotations and real clinical workflows.

Future work can broaden the set of generators and evaluators, validate safety judgments with medical experts, and extend the evaluation to multi turn settings where risk depends on longer context. It will also be important to test robustness against prompt injection and other adversarial strategies that arise in tool integrated applications. More specialized evaluation criteria for clinical subdomains, such as pediatrics or oncology, are another clear direction.

\bibliographystyle{IEEEtran}
\bibliography{Ref}

@article{chen2025mdteamgpt,
  title={Mdteamgpt: A self-evolving llm-based multi-agent framework for multi-disciplinary team medical consultation},
  author={Chen, Kai and Li, Xinfeng and Yang, Tianpei and Wang, Hewei and Dong, Wei and Gao, Yang},
  journal={arXiv preprint arXiv:2503.13856},
  year={2025}
}

@article{mukherjee2024polaris,
  title={Polaris: A safety-focused llm constellation architecture for healthcare},
  author={Mukherjee, Subhabrata and Gamble, Paul and Ausin, Markel Sanz and Kant, Neel and Aggarwal, Kriti and Manjunath, Neha and Datta, Debajyoti and Liu, Zhengliang and Ding, Jiayuan and Busacca, Sophia and others},
  journal={arXiv preprint arXiv:2403.13313},
  year={2024}
}

@inproceedings{zhao-etal-2024-llms,
    title = "Can {LLM}s Replace Clinical Doctors? Exploring Bias in Disease Diagnosis by Large Language Models",
    author = "Zhao, Yutian  and
      Wang, Huimin  and
      Liu, Yuqi  and
      Suhuang, Wu  and
      Wu, Xian  and
      Zheng, Yefeng",
    editor = "Al-Onaizan, Yaser  and
      Bansal, Mohit  and
      Chen, Yun-Nung",
    booktitle = "Findings of the Association for Computational Linguistics: EMNLP 2024",
    month = nov,
    year = "2024",
    address = "Miami, Florida, USA",
    publisher = "Association for Computational Linguistics",
    url = "https://aclanthology.org/2024.findings-emnlp.814/",
    doi = "10.18653/v1/2024.findings-emnlp.814",
    pages = "13914--13935"
}

@inproceedings{long-etal-2024-multi-expert,
    title = "Multi-expert Prompting Improves Reliability, Safety and Usefulness of Large Language Models",
    author = "Long, Do Xuan  and
      Yen, Duong Ngoc  and
      Luu, Anh Tuan  and
      Kawaguchi, Kenji  and
      Kan, Min-Yen  and
      Chen, Nancy F.",
    editor = "Al-Onaizan, Yaser  and
      Bansal, Mohit  and
      Chen, Yun-Nung",
    booktitle = "Proceedings of the 2024 Conference on Empirical Methods in Natural Language Processing",
    month = nov,
    year = "2024",
    address = "Miami, Florida, USA",
    publisher = "Association for Computational Linguistics",
    url = "https://aclanthology.org/2024.emnlp-main.1135/",
    doi = "10.18653/v1/2024.emnlp-main.1135",
    pages = "20370--20401"
}

@inproceedings{wang-etal-2024-augmenting,
    title = "Augmenting Black-box {LLM}s with Medical Textbooks for Biomedical Question Answering",
    author = "Wang, Yubo  and
      Ma, Xueguang  and Chen, Wenhu",
    editor = "Al-Onaizan, Yaser  and
      Bansal, Mohit  and
      Chen, Yun-Nung",
    booktitle = "Findings of the Association for Computational Linguistics: EMNLP 2024",
    month = nov,
    year = "2024",
    address = "Miami, Florida, USA",
    publisher = "Association for Computational Linguistics",
    url = "https://aclanthology.org/2024.findings-emnlp.95/",
    doi = "10.18653/v1/2024.findings-emnlp.95",
    pages = "1754--1770"
}

@inproceedings{liao-etal-2024-medcare,
    title = "{M}ed{C}are: Advancing Medical {LLM}s through Decoupling Clinical Alignment and Knowledge Aggregation",
    author = "Liao, Yusheng  and
      Jiang, Shuyang  and
      Chen, Zhe  and
      Wang, Yu  and
      Wang, Yanfeng",
    editor = "Al-Onaizan, Yaser  and
      Bansal, Mohit  and
      Chen, Yun-Nung",
    booktitle = "Findings of the Association for Computational Linguistics: EMNLP 2024",
    month = nov,
    year = "2024",
    address = "Miami, Florida, USA",
    publisher = "Association for Computational Linguistics",
    url = "https://aclanthology.org/2024.findings-emnlp.619/",
    doi = "10.18653/v1/2024.findings-emnlp.619",
    pages = "10562--10581"
}

@inproceedings{yu-etal-2024-cosafe,
    title = "{C}o{S}afe: Evaluating Large Language Model Safety in Multi-Turn Dialogue Coreference",
    author = "Yu, Erxin  and
      Li, Jing  and
      Liao, Ming  and
      Wang, Siqi  and
      Zuchen, Gao  and
      Mi, Fei  and
      Hong, Lanqing",
    editor = "Al-Onaizan, Yaser  and
      Bansal, Mohit  and
      Chen, Yun-Nung",
    booktitle = "Proceedings of the 2024 Conference on Empirical Methods in Natural Language Processing",
    month = nov,
    year = "2024",
    address = "Miami, Florida, USA",
    publisher = "Association for Computational Linguistics",
    url = "https://aclanthology.org/2024.emnlp-main.968/",
    doi = "10.18653/v1/2024.emnlp-main.968",
    pages = "17494--17508"
}

@article{nassiri2024recent,
  title={Recent advances in large language models for healthcare},
  author={Nassiri, Khalid and Akhloufi, Moulay A},
  journal={BioMedInformatics},
  volume={4},
  number={2},
  pages={1097--1143},
  year={2024},
  publisher={MDPI}
}

@article{luo2022biogpt,
  title={BioGPT: generative pre-trained transformer for biomedical text generation and mining},
  author={Luo, Renqian and Sun, Liai and Xia, Yingce and Qin, Tao and Zhang, Sheng and Poon, Hoifung and Liu, Tie-Yan},
  journal={Briefings in bioinformatics},
  volume={23},
  number={6},
  pages={bbac409},
  year={2022},
  publisher={Oxford University Press}
}

@article{singhal2023large,
  title={Large language models encode clinical knowledge},
  author={Singhal, Karan and Azizi, Shekoofeh and Tu, Tien-Ju and Mahdavi, S. Sara and Wei, Jason and Chung, Hyung Won and Scales, Nathan and Tanwani, Ajay and Cole, Adam and Lee, Jin and others},
  journal={Nature},
  volume={620},
  number={7972},
  pages={172--180},
  year={2023}
}

@article{onyekwelu2025emotion,
  title   = {Emotion Detection in Speech Using Lightweight and Transformer-Based Models: A Comparative and Ablation Study},
  author  = {Onyekwelu-Udoka, Lucky and Islam, Md Shafiqul and Hasan, Md Shahedul},
  journal = {arXiv preprint arXiv:2511.00402},
  year    = {2025},
  url     = {https://arxiv.org/abs/2511.00402}
}

@article{gu2021domain,
  title={Domain-specific language model pretraining for biomedical natural language processing},
  author={Gu, Yu and Tinn, Robert and Cheng, Hao and Lucas, Matthew and Usuyama, Naoto and Liu, Xiaodong and Naumann, Tristan and Gao, Jianfeng and Poon, Hoifung},
  journal={ACM Transactions on Computing for Healthcare (HEALTH)},
  volume={3},
  number={1},
  pages={1--23},
  year={2021}
}

@article{miotto2023medical,
  title={Medical AI safety: what happens when prediction fails?},
  author={Miotto, Riccardo and Wang, Fei and Jiang, Xiaoqian},
  journal={NPJ Digital Medicine},
  volume={6},
  number={1},
  pages={1--4},
  year={2023}
}

@article{yao2023tree,
  title={Tree of thoughts: Deliberate problem solving with large language models},
  author={Yao, Shinn and Zhao, Jeffrey and Yu, Dian and Zhao, Izheng and Khot, Tushar and Sabharwal, Ashish and Bosselut, Antoine and Choi, Yejin},
  journal={arXiv preprint arXiv:2305.10601},
  year={2023}
}

@article{du2023improving,
  title={Improving factual correctness in LLMs through multi-agent debate},
  author={Du, Yifan and Wang, Zhecheng and Li, Xisen and Gao, Tianyu and Zhang, Xiang and Liu, Zhiyuan},
  journal={arXiv preprint arXiv:2305.14325},
  year={2023}
}

@article{han2024medsafetybench,
  title={MedSafetyBench: Evaluating and Improving the Medical Safety of Large Language Models},
  author={Han, Tessa and Kumar, Aounon and Agarwal, Chirag and Lakkaraju, Himabindu},
  journal={arXiv preprint arXiv:2406.12345},
  year={2024}
}

@misc{ama_principles,
  title = {Principles of Medical Ethics},
  author = {{American Medical Association}},
  year = {2001},
  note = {Accessed: 2024-07-08},
  url = {https://code-medical-ethics.ama-assn.org/principles}
}

@article{liu2024survey_medllm,
  title        = {A Survey on Medical Large Language Models},
  author       = {Liu, L. and others},
  journal      = {arXiv preprint arXiv:2406.03712},
  year         = {2024}
}

@article{williams2024clinical_recommendations,
  title        = {Evaluating Large Language Models for Clinical Recommendations from Emergency Department Notes},
  author       = {Williams, C. Y. K. and others},
  journal      = {Nature Communications},
  year         = {2024},
  note         = {Article: s41467-024-52415-1}
}

@article{aydin2024patient_education_review,
  title        = {Large Language Models in Patient Education: A Scoping Review},
  author       = {Aydin, S. and others},
  journal      = {Frontiers in Medicine},
  year         = {2024},
  doi          = {10.3389/fmed.2024.1477898}
}

@article{birkun2024heart_attack_chatbot,
  title        = {Large Language Model-based Chatbot as a Source of Advice on First Aid in Heart Attack},
  author       = {Birkun, A. A. and others},
  journal      = {American Journal of Emergency Medicine},
  year         = {2024}
}

@article{birkun2023cpr_chatbot,
  title        = {Large Language Model (LLM)-Powered Chatbots Fail to Reliably Guide Cardiopulmonary Resuscitation},
  author       = {Birkun, A. A. and others},
  journal      = {Resuscitation},
  year         = {2023}
}

@article{ouyang2022instructgpt,
  title        = {Training Language Models to Follow Instructions with Human Feedback},
  author       = {Ouyang, Long and Wu, Jeffrey and Jiang, Xu and Almeida, Diogo and Wainwright, Carroll and Mishkin, Pamela and others},
  journal      = {Advances in Neural Information Processing Systems},
  year         = {2022},
  note         = {arXiv:2203.02155}
}

@article{bai2022constitutional_ai,
  title        = {Constitutional AI: Harmlessness from AI Feedback},
  author       = {Bai, Yuntao and Kadavath, Saurav and Kundu, Sandipan and Askell, Amanda and Kernion, Jackson and Jones, Andy and others},
  journal      = {arXiv preprint arXiv:2212.08073},
  year         = {2022}
}

@article{liu2023prompt_injection,
  title        = {Prompt Injection Attacks against LLM-Integrated Applications},
  author       = {Liu, Yi and others},
  journal      = {arXiv preprint arXiv:2306.05499},
  year         = {2023}
}

@misc{owasp_llm_top10,
  title        = {OWASP Top 10 for Large Language Model Applications},
  author       = {{OWASP Foundation}},
  howpublished = {\url{https://owasp.org/www-project-top-10-for-large-language-model-applications/}},
  year         = {2024},
  note         = {Accessed: 2025-12-25}
}

@article{lee2020biobert,
  title   = {BioBERT: a pre-trained biomedical language representation model for biomedical text mining},
  author  = {Lee, Jinhyuk and Yoon, Wonjin and Kim, Sungdong and Kim, Donghyeon and Kim, Sunkyu and So, Chan Ho and Kang, Jaewoo},
  journal = {Bioinformatics},
  volume  = {36},
  number  = {4},
  pages   = {1234--1240},
  year    = {2020},
  doi     = {10.1093/bioinformatics/btz682}
}

@article{yang2022gatortron,
  title   = {A large language model for electronic health records},
  author  = {Yang, Xi and Chen, Aokun and PourNejatian, Nima and Shin, Hoo Chang and Smith, Kaleb E. and Parisien, Christopher and Compas, Colin and Martin, Cheryl and Costa, Anthony B. and Flores, Mona G. and others},
  journal = {npj Digital Medicine},
  volume  = {5},
  pages   = {194},
  year    = {2022},
  doi     = {10.1038/s41746-022-00742-2}
}

@inproceedings{lu-etal-2022-clinicalt5,
  title     = {ClinicalT5: A Generative Language Model for Clinical Text},
  author    = {Lu, Qiuhao and Dou, Dejing and Nguyen, Thien},
  booktitle = {Findings of the Association for Computational Linguistics: EMNLP 2022},
  pages     = {5436--5443},
  year      = {2022},
  doi       = {10.18653/v1/2022.findings-emnlp.398}
}

@article{chen2023meditron,
  title   = {MEDITRON-70B: Scaling Medical Pretraining for Large Language Models},
  author  = {Chen, Zeming and Hern{\'a}ndez Cano, Alejandro and Romanou, Angelika and Bonnet, Antoine and Matoba, Kyle and Salvi, Francesco and Pagliardini, Matteo and Fan, Simin and K{\"o}pf, Andreas and Mohtashami, Amirkeivan and others},
  journal = {arXiv preprint arXiv:2311.16079},
  year    = {2023}
}

@article{madaan2023selfrefine,
  title   = {Self-Refine: Iterative Refinement with Self-Feedback},
  author  = {Madaan, Aman and Tandon, Niket and Gupta, Prakhar and Hallinan, Skyler and Gao, Luyu and Wiegreffe, Sarah and Alon, Uri and Dziri, Nouha and Prabhumoye, Shrimai and Yang, Yiming and others},
  journal = {arXiv preprint arXiv:2303.17651},
  year    = {2023}
}

@article{shinn2023reflexion,
  title   = {Reflexion: Language Agents with Verbal Reinforcement Learning},
  author  = {Shinn, Noah and Cassano, Federico and Berman, Edward and Gopinath, Ashwin and Narasimhan, Karthik and Yao, Shunyu},
  journal = {arXiv preprint arXiv:2303.11366},
  year    = {2023}
}

\end{document}